\documentclass[pdflatex,sn-mathphys-num]{sn-jnl}% Math and Physical Sciences Numbered Reference Style
\usepackage{graphicx}%
\usepackage{multirow}%
\usepackage{amsmath,amssymb,amsfonts}%
\usepackage{amsthm}%
\usepackage{mathrsfs}%
\usepackage[title]{appendix}%
\usepackage{xcolor}%
\usepackage{textcomp}%
\usepackage{manyfoot}%
\usepackage{booktabs}%
\usepackage{algorithm}%
\usepackage{algorithmicx}%
\usepackage{algpseudocode}%
\usepackage{listings}%
\usepackage{svg}%
\usepackage{subcaption}%
\usepackage{titlesec}%
\usepackage{pdflscape}%
\usepackage{hyperref}%
\usepackage{makecell}%
\theoremstyle{thmstyleone}%
\theoremstyle{thmstyletwo}%

\theoremstyle{thmstylethree}%

\titleformat{\paragraph}
{\normalfont\normalsize\bfseries}{\theparagraph}{1em}{}
\titlespacing*{\paragraph}
{0pt}{3.25ex plus 1ex minus .2ex}{1.5ex plus .2ex}

\begin{document}

\title[MinkUNeXt-SI]{MinkUNeXt-SI: Improving point cloud-based place recognition including spherical coordinates and LiDAR intensity}


\author*[1]{\fnm{Judith} \sur{Vilella-Cantos}}\email{jvilella@umh.es}

\author[1]{\fnm{Juan José} \sur{Cabrera}}\email{juan.cabreram@umh.es}

\author[1]{\fnm{Luis} \sur{Payá}}\email{lpaya@umh.es}

\author[1]{\fnm{Mónica} \sur{Ballesta}}\email{m.ballesta@umh.es}

\author[1]{\fnm{David} \sur{Valiente}}\email{dvaliente@umh.es}

\affil*[1]{\orgdiv{University Institute for Engineering Research}, \orgname{Miguel Hernández University}, \orgaddress{\street{Avda. de la Universidad s/n, Edificio Innova}, \city{Elche}, \postcode{03202}, \state{Alicante}, \country{Spain}}}

%\affil[2]{\orgdiv{Department}, \orgname{Organization}, \orgaddress{\street{Street}, \city{City}, \postcode{10587}, \state{State}, \country{Country}}}

%\affil[3]{\orgdiv{Department}, \orgname{Organization}, \orgaddress{\street{Street}, \city{City}, \postcode{610101}, \state{State}, \country{Country}}}


\abstract{In autonomous navigation systems, the solution of the place recognition problem is crucial for their safe functioning. But this is not a trivial solution, since it must be accurate regardless of any changes in the scene, such as seasonal changes and different weather conditions, and it must be generalizable to other environments. This paper presents our method, MinkUNeXt-SI, which, starting from a LiDAR point cloud, preprocesses the input data to obtain its spherical coordinates and intensity values normalized within a range of 0 to 1 for each point, and it produces a robust place recognition descriptor. To that end, a deep learning approach that combines Minkowski convolutions and a U-net architecture with skip connections is used. The results of MinkUNeXt-SI demonstrate that this method reaches and surpasses state-of-the-art performance while it also generalizes satisfactorily to other datasets. Additionally, we showcase the capture of a custom dataset and its use in evaluating our solution, which also achieves outstanding results. Both the code of our solution and the runs of our dataset are publicly available for reproducibility purposes\footnote{\url{https://github.com/JudithV/MinkUNeXt-SI}}.}




\maketitle

\section{Introduction}\label{sec1}
Place recognition is a current challenge, since many systems—particularly autonomous navigation systems—depend on solving it for proper operation. This problem consists of accurately determining a system’s location within a map by establishing a precise correspondence between the environment captured by sensors at a given moment and other previous version of the environment stored in a database. Besides, a reliable place recognition solution is also crucial for developing an effective SLAM (Simultaneous Localization and Mapping) system, which is indispensable for any mobile robotics application. This is because loop closure enables real-time error correction, ensuring the system's correct localization in space.

The place recognition problem has been approached from various perspectives, using input data captured by different types of sensors. The most extensively studied and pioneering sensor in place recognition tasks is the visual sensor \cite{knopp2010}. For this task, cameras of all kinds have been used, employing a wide range of perspectives \cite{lee2019, schleiss2022}. Over time, place recognition solutions have been adapted and developed natively to other types of sensors, such as RADAR \cite{gadd2024, peng2024}, Light Detection and Ranging (LiDAR) \cite{uy2018}, and SONAR \cite{santos2019, gaspar2023}. This work concentrates on the use of sensory data acquired by LiDAR, because while the presence on seasonal changes remains a challenge in place recognition tasks due to changes in the geometric structures that make up the environment (per example, the lack or abundance of leaves on a tree), this type of sensors are mostly invariant to changes in lighting and weather conditions compared to the traditional approach of relying only in visual information.

Beyond input data, it is also relevant to consider additional information that enhances descriptor accuracy. Traditionally, place recognition methods rely solely on sensor data as the only input \cite{komorowski2021}. However, other studies demonstrate that incorporating additional information into the input and preprocessing data—such as intensity and reflectance in point clouds \cite{viswanath2024} or thermal information in images \cite{deng2021, heredia2025}—yields better results compared to using only raw sensor data in deep learning tasks like segmentation. Currently, efforts are being made to extend these implementations to place recognition tasks \cite{zywanowski2022}. Considering the evidence that additional information improves the performance of deep learning algorithms for place recognition, another approach to use a more accurate representation of the environment by using a different coordinate representation has also been explored in recent years, showing that this representation provides an accurate description of the geometric elements captured by the LiDAR sensor \cite{rao2019, chang2020}.

In order to provide a robust solution for localization in mobile robotics, we combine the two approaches presented in the previous paragraph (spherical representation and inclusion of the intensity value of the points) in order to merge the most innovative approaches taken in this field in recent years. The present work aims to contribute an innovative advancement by combining state-of-the-art techniques that yield superior results: convolutional neural networks, a deep U-Net architecture, and 4D Minkowski convolutions applied to polar coordinate spaces, enriched with additional information such as point intensity. The goal of our work is to demonstrate that using the spherical representation of coordinates and the LiDAR intensity value are powerful additions to a deep learning technique. Specifically, the preprocessing of these input values improves an already well-performing place recognition neural network: MinkUNeXt \cite{cabrera2024}. This method produces a robust descriptor of the environment against seasonal and dynamic variations.

The main contributions of this work are:
\begin{itemize}
    \item MinkUNeXt preprocessing modifications \cite{cabrera2024}, setting the network input as intensity information and spherical coordinates, achieving and overpassing the state-of-the-art in place recognition.
    \item Benchmarking of our method with different datasets containing seasonal, environmental and other dynamic changes, obtaining results that surpass the current state-of-the-art.
    \item A custom-collected dataset for evaluating the proposed solution.
\end{itemize}

In this work, the different sections are distributed in the following way. Section \ref{sec2} reviews the state-of-the-art for the topics in the context of the current paper. Section \ref{sec3} presents the different methods used for data processing as well as the backbone of our method. Section \ref{sec4} elaborates on the setup of the experiments, including information on all the datasets involved in this work, the hyperparameter configuration of the network, and the clustering obtained with our MinkUNeXt-SI. Section \ref{sec5} presents the results obtained by our MinkUNeXt-SI in different scenarios and compares them with other place recognition methods. Finally, Section \ref{sec6} draws some conclusions about MinkUNeXt-SI and proposes future work in this line of research.

\section{Related works}\label{sec2}

\subsection{Image retrieval for place recognition}\label{subsec21}
The use of image representation in deep learning problems has been extensively researched over the last decades. This is because image-based sensors have always been considered the gold standard for vision systems in robotics, as they can capture increasingly high-resolution representations of the surrounding environment.

In particular, the place recognition problem is one of the most studied topics in mobile robotics, as solving localization is crucial for the correct functioning of these systems. A reliable solution must enable the system to recognize its location while remaining invariant to changes in lighting, seasons, or adverse weather conditions \cite{cebollada2023}. Several applications benefit from addressing this challenge, including autonomous vehicles \cite{juneja2023} and drones \cite{schleiss2022}, which require a robust and consistent recognition system to function properly\cite{masone2021}.

The use of visual descriptors to describe the environment based on image data has been extensively explored in recent decades \cite{bay2008, lowe2004}. The first and most well-known approach to addressing the place recognition problem using images is NetVLAD \cite{arandjelovic2016}, which proposes a solution through a convolutional neural network combined with a VLAD pooling layer. Other contributions, such as \cite{galvez2012}, suggest using the Bag-of-Words (BoW) model for fast recognition from images, employing FAST+BRIEF to extract features from them.

Recent solutions applied to the place recognition problem include studies like \cite{le2020}, which focus on large-scale localization (e.g., city-wide localization) and emphasize the need for diverse training datasets to achieve reliable results. Other approaches aim to enrich image data for these applications by combining geolocation methods with deep learning techniques \cite{pramanick2022, clark2023,yang2021}. 

Additionally, recent studies are exploring the potential of Vision-Language Models \cite{waheed2025, qi2017}, which aim to process image data similarly to words, providing a more precise description than traditional approaches. However, this line of research is still in an exploratory and refinement phase for most applications.

One of the main challenges of localization using visual image data is that the system becomes highly susceptible to weather, lighting, and seasonal changes \cite{juneja2023}. A single location can appear drastically different under varying environmental conditions. Even for the human eye, determining whether two images correspond to the same place can be difficult when such variations are pronounced—for example, when an environment is covered in snow.

\subsection{LiDAR representation in deep learning}\label{subsec22}
Over time, researchers began considering the use of geometric structures to represent the environment instead of conventional vision systems \cite{maturana2015}. The advantage of these 3D reconstructions is that they offer greater invariance to various changes compared to images, as they are not affected by lighting variations. The LiDAR sensor is the most well-known tool for obtaining point clouds of the environment. This sensor reconstructs the scene by emitting infrared pulses that bounce off the surfaces of objects, and based on the time it takes for the pulse to return to the sensor, the position of each point in the environment is calculated. LiDAR sensors come in various resolutions, depending on the needs and budget of each project.

Due to the robustness of this type of data, PointNetVLAD \cite{uy2018} was a pioneering approach in addressing the place recognition problem by introducing 3D structures as input to convolutional neural networks instead of traditional image-based methods. In this work, Uy and Lee proposed using PointNet \cite{qi2017} to extract features from point clouds and feed that information into NetVLAD.

Following this work, multiple approaches were explored to improve the effectiveness of PointNetVLAD, which struggled to capture complex structures in point clouds due to PointNet’s inability to detect local features. TransLoc3D \cite{xu2021} introduced Adaptive Receptive Fields to remove noise and handle variable-sized point clouds; LPD-Net \cite{liu2019} incorporated local features to better learn the structure of point clouds; PTC-Net \cite{chen2023} applied transformers to neighborhoods of multiple point clouds to enhance robustness; LPS-Net \cite{liu2024} introduced the use of BPU to differentiate multi-scale features while maintaining a lightweight model; and SOE-Net \cite{xia2021} integrated orientation encoding for each point combined with self-attention mechanisms.

A class of convolutional neural networks gaining popularity in the field of place recognition using point clouds is those utilizing Minkowski convolutions \cite{choy2019}. These sparse convolutions are more efficient when working with unstructured data, such as point clouds. Notable works using Minkowski convolutions for this application. MinkLoc3D \cite{komorowski2021} employs a feature pyramid network architecture to emphasize local features; MinkLoc3Dv2 \cite{komorowski2022} improves the original network by incorporating the Truncated Smooth AP loss function with large batch sizes; TE-NeXt \cite{santo2025} introduces a labeling system to distinguish between traversable and non-traversable terrain, enhancing how this distinction is really useful for mobile robotics in irregular environments \cite{santo2025ground}; and MinkUNeXt \cite{cabrera2024} incorporates a U-Net structure with skip connections to capture both local and global features. This last descriptor of the environment has proven to be highly effective in the place recognition task, surpassing the state-of-the-art values for the two protocols introduced in PointNetVLAD \cite{uy2018}, baseline and refined, which consist in training the descriptor with urban environments, long-term datasets (the Oxford RobotCar Dataset and an in-house dataset).

\subsection{Image and LiDAR fusion approaches}\label{subsec23}
In the search for the optimal solution to the place recognition problem, a widely adopted technique is sensor fusion as an improvement over using a single type of input data. Studies such as \cite{zywanowski2020} and \cite{komorowski2021minkloc++} demonstrate that combining camera and LiDAR data yields better results across all evaluated scenarios.

However, using both camera and LiDAR data simultaneously is not trivial. Processing large volumes of data in different formats is computationally demanding, and integrating additional sensors increases the system's cost and complexity. Therefore, a more affordable alternative is to combine LiDAR data with additional information that enriches the features provided by each point cloud. Specifically, incorporating the intensity values of each point has been shown in various studies \cite{guo2019} to improve performance compared to excluding it, as it provides additional information about the materials composing the geometric structures present in the environment \cite{park2021}.

Another key aspect to consider when representing an object's coordinates in three-dimensional space is the choice between Cartesian and polar (also known as spherical) coordinates. The traditional approach used in most studies \cite{qi2017_plusplus, xu2018, guerrero2018} relies on Cartesian representation, as it is the standard reference system for sensors and it is commonly used in voxelization-based processing techniques. However, other studies \cite{lei2018, jung2025, zywanowski2022} propose spherical representation as a more accurate way to depict 3D structures, as this format better captures the positioning and relationships between objects detected by the sensor. While fusing different sensors leads to an expensive, complicated and computationally demanding solution, the approach of combining different information captured by the same sensor can be a much more intuitive and efficient solution. This approach has been tested with different features, but one of the most common is the inclusion of intensity or reflectivity information in point clouds, as it provides an accurate metric for distinguishing the materials that form the elements present in the geometry captured by LiDAR \cite{wang2020}.

\section{Methodology}\label{sec3}
\begin{figure}
    \centering
    \includegraphics[width=1.1\linewidth]{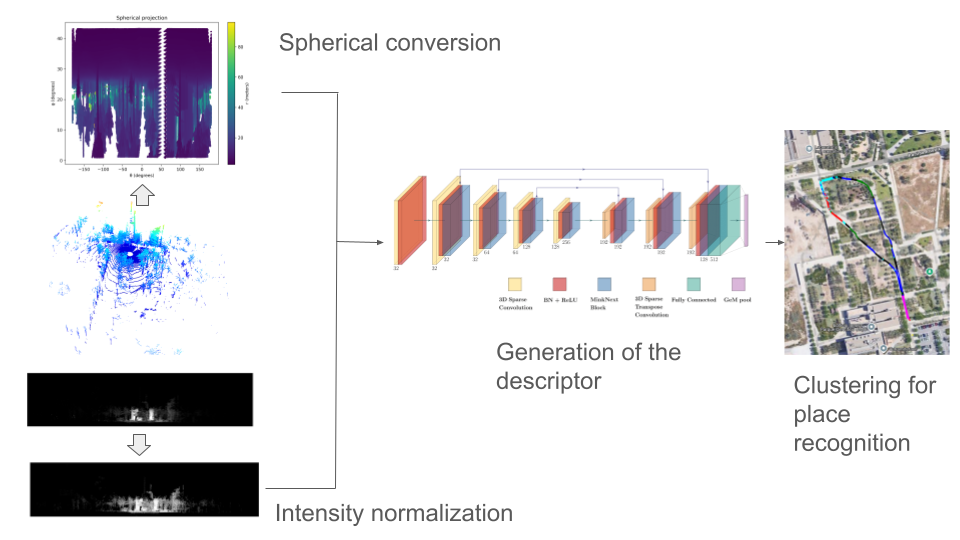}
    \caption{MinkUNeXt-SI's workflow. First, the input point cloud is processed from Cartesian to spherical coordinates, plus applying histogram equalization to the intensity channel. This processed information then feeds our MinKUNeXt-SI, resulting in a robust descriptor of the environment for place recognition.}
    \label{diagram}
\end{figure}

Diagram shown in Figure \ref{diagram} presents the functioning workflow of MinkUNeXt-SI. Our method takes as input a LiDAR point cloud, an unordered 3D structure defined as $\mathcal{P}$ along with a feature vector corresponding to the intensity value associated with each point in the point cloud. 

\[
\mathcal{P} = \left\{ \mathbf{p}_i = (x_i, y_i, z_i) \in \mathbb{R}^3 \mid i = 1, 2, \dots, N \right\}
\]

Both components are quantized into sparse tensors and processed through the network layers. The U-Net architecture with skip connections ensures the fusion of local and global features, resulting in a robust descriptor for place recognition, independent of seasonal or dynamic changes, and capable of generalizing effectively to other datasets.

The preprocessing of the data by our method includes the transformation from Cartesian to spherical coordinates and the normalization of intensity values in datasets that do not provide this parameter in a range of 0 to 1. The normalization of intensity values is performed using the histogram equalization method, a common approach to enhancing contrast between pixel values in an image \cite{moon2017}. This results in a more evenly distributed range of values rather than having them concentrated in distinct peaks. It is a widely used data preprocessing technique in computer vision applications. In our case, we apply this approach to the intensity values of each point in a point cloud, instead of image pixels, in order to ensure that the inclusion of intensity information translates into a meaningful input for the task of place recognition.

\subsection{Spherical transformation}\label{subsec31}
While Minkowski convolutions operate in Cartesian space, the spherical representation of 3D point cloud coordinates obtained from a LiDAR sensor provides a more precise depiction of the general structure of the environment and captures object relationships more accurately. Given the Cartesian coordinates $(x_i, y_i, z_i)$ of a point $i$, its polar coordinates $(r_i, \theta_i, \phi_i)$ can be computed using the following equations:

\begin{align}
r_i &=  \sqrt{x^2_i + y^2_i +z^2_i}\ \\
\theta_i &= arctan2(y_i, x_i) \\
\phi_i &= arccos(z_i, r_i)
\label{eq_spherical}
\end{align}

Here, $r_i$ represents the distance between the point in the point cloud and the reference of the sensor, while $\phi_i$ and $\theta_i$ correspond to the vertical and horizontal angles in the scan, respectively.

When working with different datasets, each acquired using a distinct LiDAR sensor, it is essential to consider the specifications of each sensor to ensure the correct spherical transformation. This includes taking into account the Field of View (FoV) range for the given sensor and applying the necessary transformations to compute $\theta_i$ accordingly.

\begin{figure}[h]
    \centering
    \includegraphics[width=0.6\linewidth]{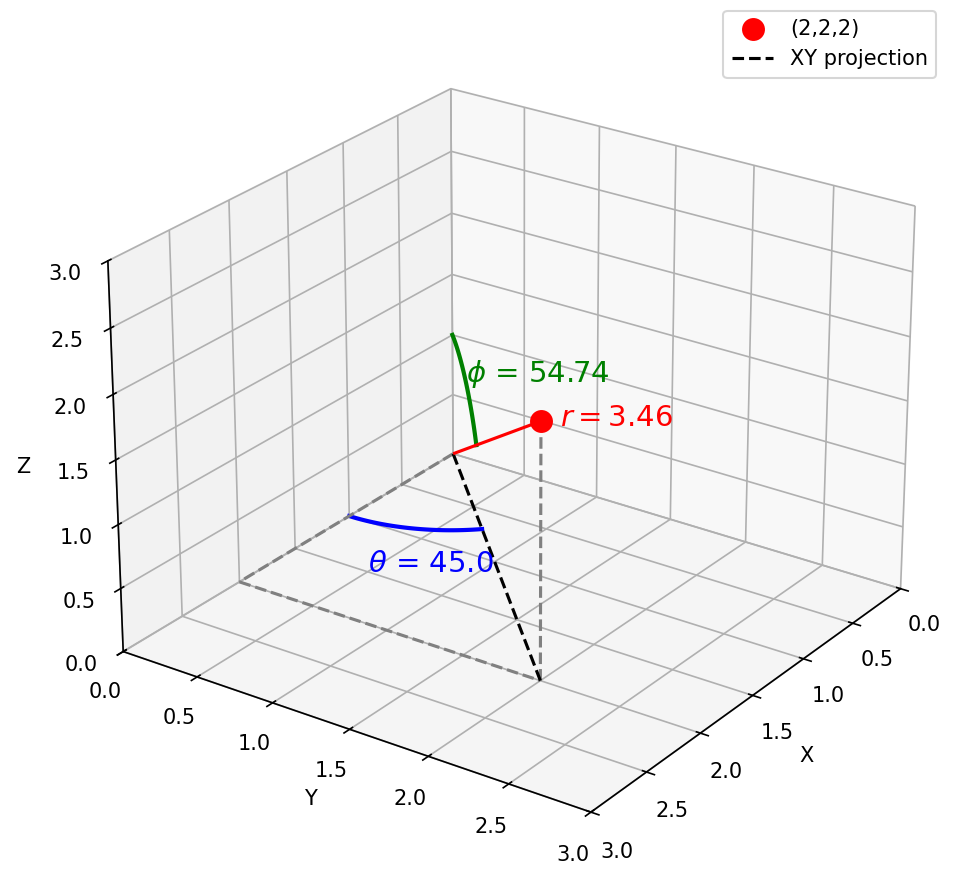}
    \caption{Representation of a sample point in the Cartesian space $P = (2, 2, 2)$ with its spherical conversion resulting in $P = (3.46, 45, 54.7)$.}
    \label{spherical-conversion}
\end{figure}

For clarity, Figure \ref{spherical-conversion} presents the representation of the same point in the Cartesian and spherical space.

\subsection{Equalization of the intensity}\label{subsec32}

\begin{figure}[h]
    \centering
    \includegraphics[width=\textwidth]{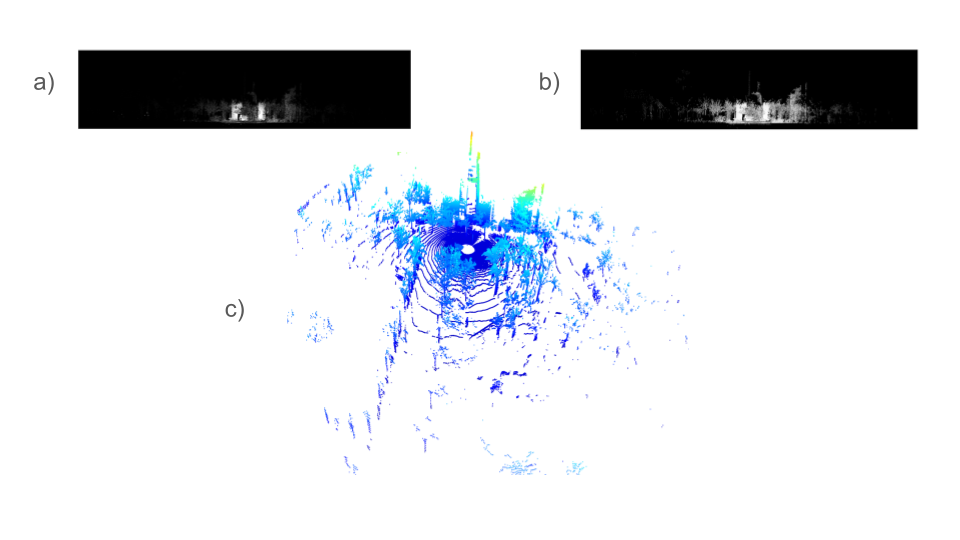}
    \caption{Comparison between the projected image from the same point cloud colored with the intensity values: (a) not normalized; (b) normalized; (c) original point cloud.}
    \label{equalization}
\end{figure}

The fact that intensity values can vary significantly (depending on the sensor, intensity may range from 0 to 255, 3000, or even 4000) can make this feature less descriptive when used as additional input to a neural network alongside 3D structures. Therefore, in our solution, intensity values are normalized between 0 and 1 before being used as input to the neural network in the form of a feature vector. Figure \ref{equalization} illustrates the differences between normalized intensity values and non-normalized values for the same point cloud. Notice how the normalized intensity provides more information about the structures that conform the environment.

While the datasets used for training already had this value normalized for each scan, in the ARVC (Automation, Robotics and Computer Vision lab) and NCLT datasets, normalization was performed using histogram equalization.

The histogram was equalized using the \textit{scikit} library in Python, which processes intensity as follows. Given the probability of occurrence of an intensity level obtained using the formula:

\begin{equation}
p(i) = \frac{h(i)}{N}
\label{eqIntensity1}
\end{equation}

where $h(i)$ is the number of times a given intensity value appears in the scan, and $N$ is the total number of points, we then compute the cumulative histogram using:

\begin{equation}
C(i) = \sum_{j=0}^{i}{p(i)}
\label{eqIntensity2}
\end{equation}

Finally, the intensity values are transformed using the following equation:

\begin{equation}
s(i) = \frac{C(i) - C_{min}}{N - C_{min}}\times(L - 1)
\label{eqIntensity3}
\end{equation}

where $C_{min}$ is the smallest cumulative intensity value greater than one, and $L$ is the total number of possible intensity values.

\section{Experiments}\label{sec4}
\subsection{Datasets}\label{subsec41}
The datasets used for training our method are the Oxford RobotCar Dataset \cite{maddern2017} and the University of Sydney Campus Dataset \cite{zhou2020}. To evaluate the descriptor’s performance, we use the North Campus Long-Term Dataset from the University of Michigan \cite{cavelaris2016}, KITTI \cite{geiger2013}, and a custom-collected dataset, the ARVC Dataset \cite{arvc}.

The choice of these datasets was made for several reasons. First, the selection of the Oxford RobotCar dataset and the USyd dataset was made to obtain results that allow us to compare with the state-of-the-art in this question, specifically with \cite{zywanowski2022}. On the other hand, the selection of the validation datasets was made taking into account two of the most used datasets in the place recognition task (KITTI and NCLT) and adding our own collected data. These collections have in common that they were all collected in an urban environment (near different university campuses) and in long-term collection sessions, with the exception of the KITTI dataset, which does not correspond to long-term conditions. This means that the aforementioned datasets contain seasonal and environmental changes, with the exception of KITTI. In this way, we can test the robustness of MinkUNeXt-SI not only in different contexts, but also against all kinds of changes in the scene.
\subsubsection{Training datasets}\label{subsubsec411}
\paragraph{Oxford RobotCar Dataset}\label{subsubsubsec4111}
This dataset consists of multiple routes recorded twice a week throughout the city of Oxford over a period of more than a year, covering a total distance of more than 1000 km \cite{maddern2017}. It captures a wide range of environmental variations, including seasonal changes, different weather conditions, and dynamic elements such as pedestrians and construction work.

The dataset provides both image and 3D data. The 3D data, which is relevant to our method, is obtained using two SICK LMS-151 2D LiDAR sensors mounted on the vehicle that records data from all sensors along the route. Each point cloud consists of 4096 points, and data processing (including intensity normalization) follows the same procedure as in \cite{zywanowski2022}.

\paragraph{USyd}\label{subsubsubsec4112}
This dataset was collected by recording routes weekly for 60 consecutive weeks in an urban environment (University of Sydney) over a period of 1.5 years \cite{zhou2020}. The LiDAR sensor used to acquire the point clouds is a Velodyne Puck VLP-16, which captures data using 16 channels, generating up to 25,000 points per cloud. The preprocessing applied to this dataset follows the same procedure as in \cite{zywanowski2022}, meaning that no points were removed before processing. Additionally, the dataset was split into training and validation sets following the same partitioning strategy as in \cite{zywanowski2022}, with test areas selected as arbitrary 100x100 m regions. The dataset includes a total of 19,138 scans for training and 8,797 for testing.

\subsubsection{Generalization datasets}\label{subsubsec412}
\paragraph{KITTI}\label{subsubsubsec4121}
The KITTI dataset \cite{geiger2013} is a long-term urban environment dataset that was recorded following different routes through the city of Karlsruhe. This dataset contains data from several sensors: cameras, LiDAR, GPS, and also contains several annotated runs. Many dynamic changes have been captured by this dataset, as it was recorded following long routes through the city of Karlsruhe, so it captures both crowded and quiet areas. However, since this dataset was only collected over a two-month period, it does not allow for the capture of seasonal changes. For this reason, this dataset is very useful for the generalization task, but not so much for training.

To evaluate the generalization capability of the descriptor obtained using our method, we use the first 170 seconds of sequence 00 from the KITTI dataset, following the approach in \cite{pan2021}. The LiDAR sensor used to capture the point clouds in this dataset is a Velodyne HDL-64E, which features 64 channels and achieves a resolution of over 1.3 million points per second.

\paragraph{NCLT}\label{subsubsubsec4122}
The North Campus Long-Term dataset \cite{cavelaris2016} was collected in 2012 and 2013, following various routes across the University of Michigan campus over a 15-month period, with a total of 27 routes. Similarly to the previous dataset, the NCLT dataset is used to assess the descriptor's generalization capability, as it was collected in an urban environment over a period of more than a year, making it comparable to the datasets used for training. Data from the first three routes of the dataset are used for evaluation. The query and validation sets were created by selecting arbitrary 100x100 meter areas, following the same procedure applied to the training datasets. The LiDAR sensor used in this dataset is a Velodyne HDL-32E, capable of capturing up to 695,000 points per second.
\clearpage

\begin{figure}[h]
    \centering
    \begin{subfigure}
        [b]{0.25\textwidth}
        \includegraphics[width=\textwidth]{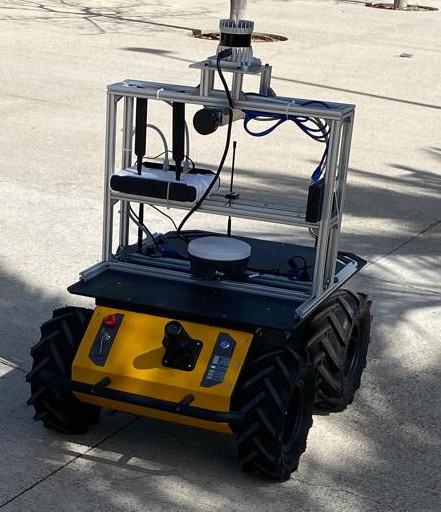}
        \caption{}
        \label{setup}
    \end{subfigure}
    \begin{subfigure}
        [b]{0.5\textwidth}
        \includegraphics[width=\textwidth]{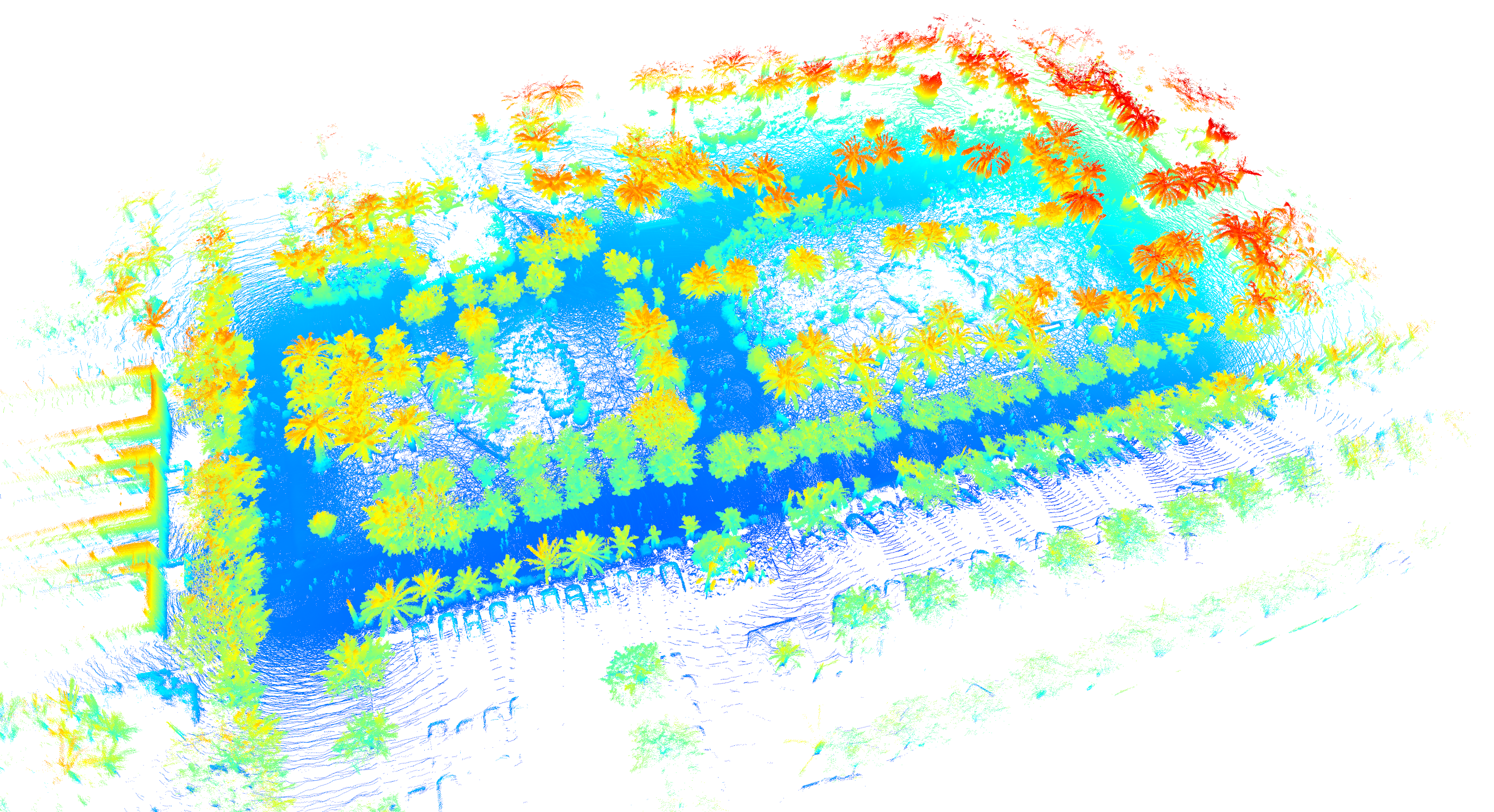}
        \caption{}
        \label{arvc-lidarmap}
    \end{subfigure}
    \caption{Some of the characteristics of the ARVC dataset. (a) Robotic setup, the robotic platform is a Husky UGV and the mounted LiDAR sensor model is an Ouster OS1-128 LiDAR. (b) Map of a run on the Miguel Hernández University campus, constructed from LiDAR scans.}
    \label{arvc}
\end{figure}

\paragraph{ARVC}\label{subsubsubsec4123}
The ARVC dataset \cite{arvc} was collected at the Miguel Hernández University of Elche over a year, recording at least 3 routes per season, covering several short routes across the campus. This dataset contains both indoor and outdoor recordings, although in this work we focus only on outdoor place recognition, and it is under continuous development. The setup for this collection consists of a Husky UGV platform, an Ouster OS1-128 LiDAR sensor, GPS-RTK (Figure \ref{setup}). 

\begin{figure}[t]
    \centering
    \begin{subfigure}
        [b]{0.4\textwidth}
        \includegraphics[width=\textwidth]{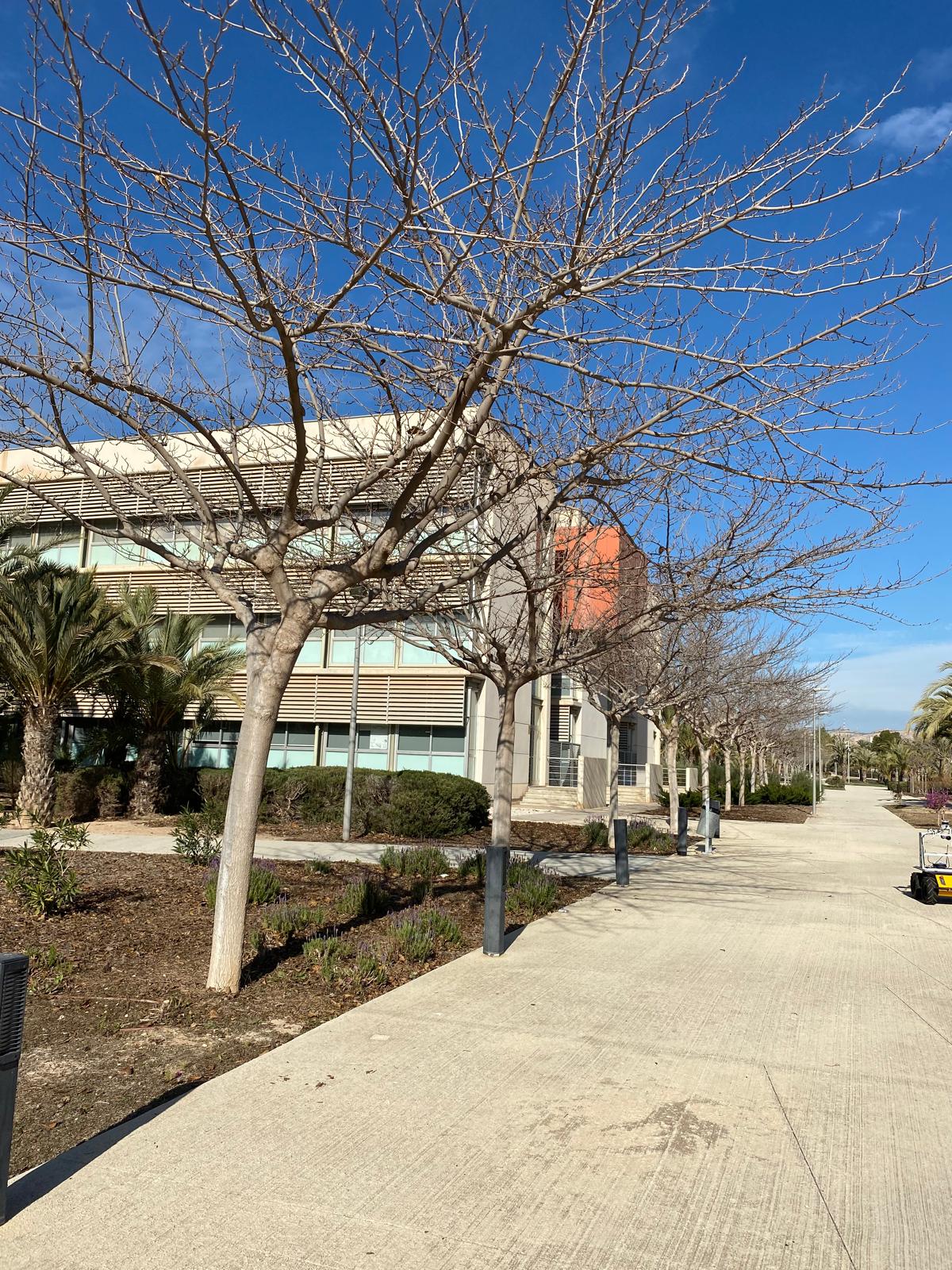}
        \caption{Image of a place captured by our ARVC dataset in winter.}
        \label{arvc-winter}
    \end{subfigure}
    \begin{subfigure}
        [b]{0.4\textwidth}
        \includegraphics[width=\textwidth]{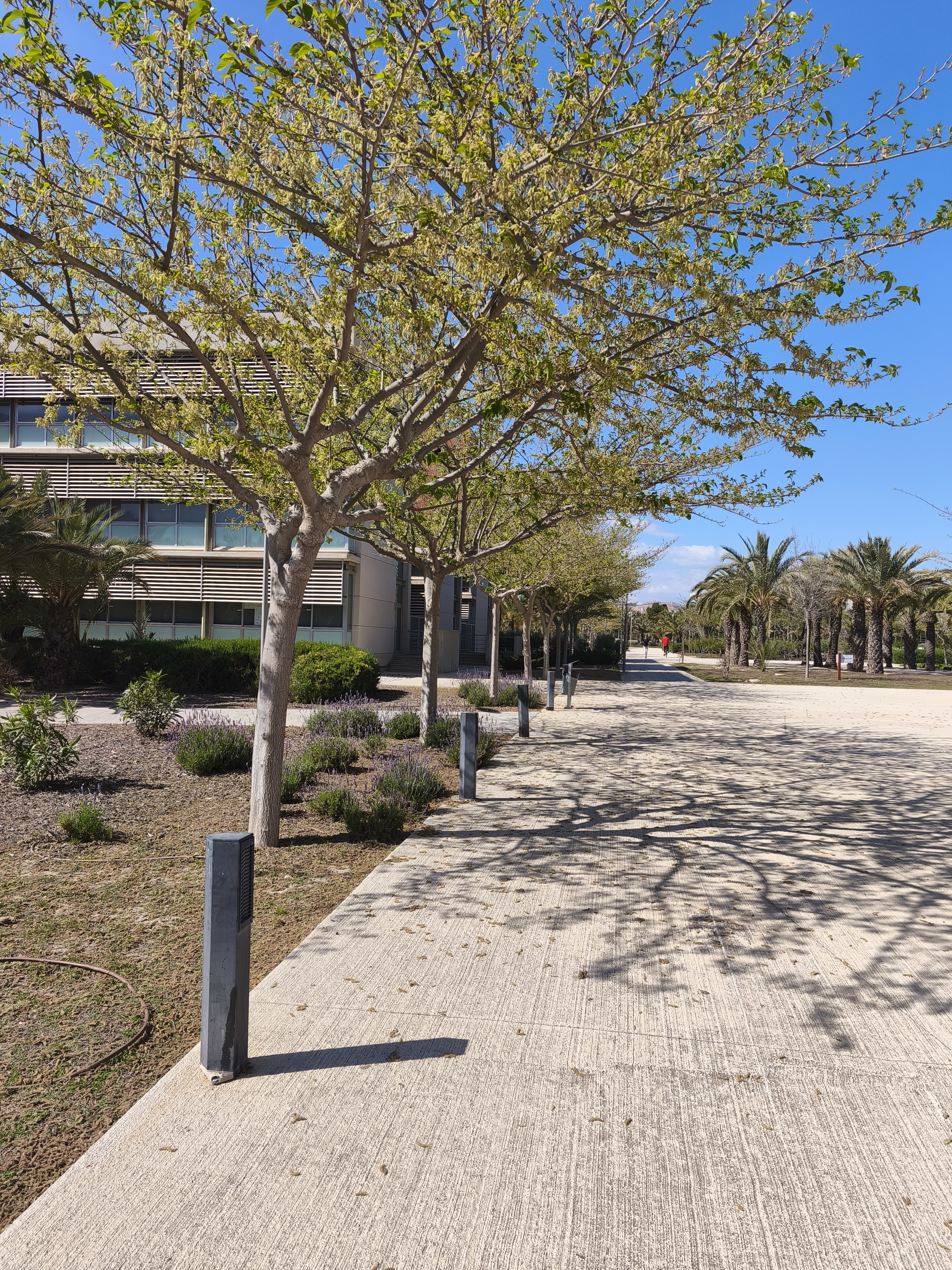}
        \caption{Image of the same location captured by our ARVC dataset in spring.}
        \label{arvc-spring}
    \end{subfigure}
    \caption{Comparison between the same location captured by our ARVC dataset in different seasons: (a) winter; (b) spring.}
    \label{arvc-seasons}
\end{figure}

Our dataset captures seasonal variations and the presence of dynamic elements in the environment, such as pedestrians, vehicles, and dogs. Figure \ref{arvc-seasons} shows the significant change between the same location taken in different seasons.

Figure \ref{arvc-lidarmap} shows a map of one of the routes from this dataset, constructed using all LiDAR scans recorded for that route. 

\begin{figure}[t]
    \centering
    \begin{subfigure}
        [b]{0.4\textwidth}
        \includegraphics[height=6cm,width=\textwidth]{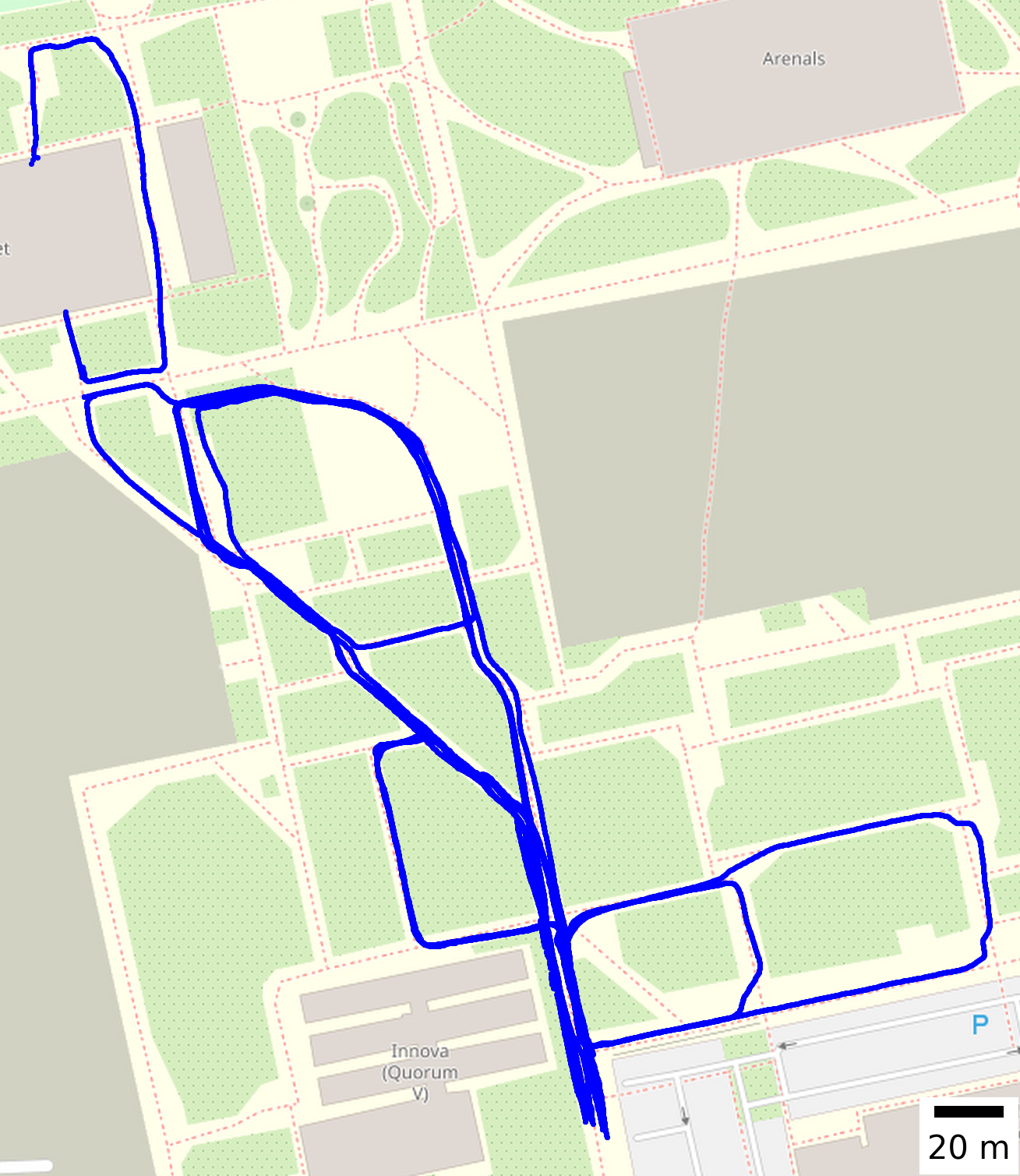}
        \caption{All routes taken by the ARVC dataset.}
        \label{arvc-all}
    \end{subfigure}
    \begin{subfigure}
        [b]{0.3\textwidth}
        \includegraphics[height=6cm,width=\textwidth]{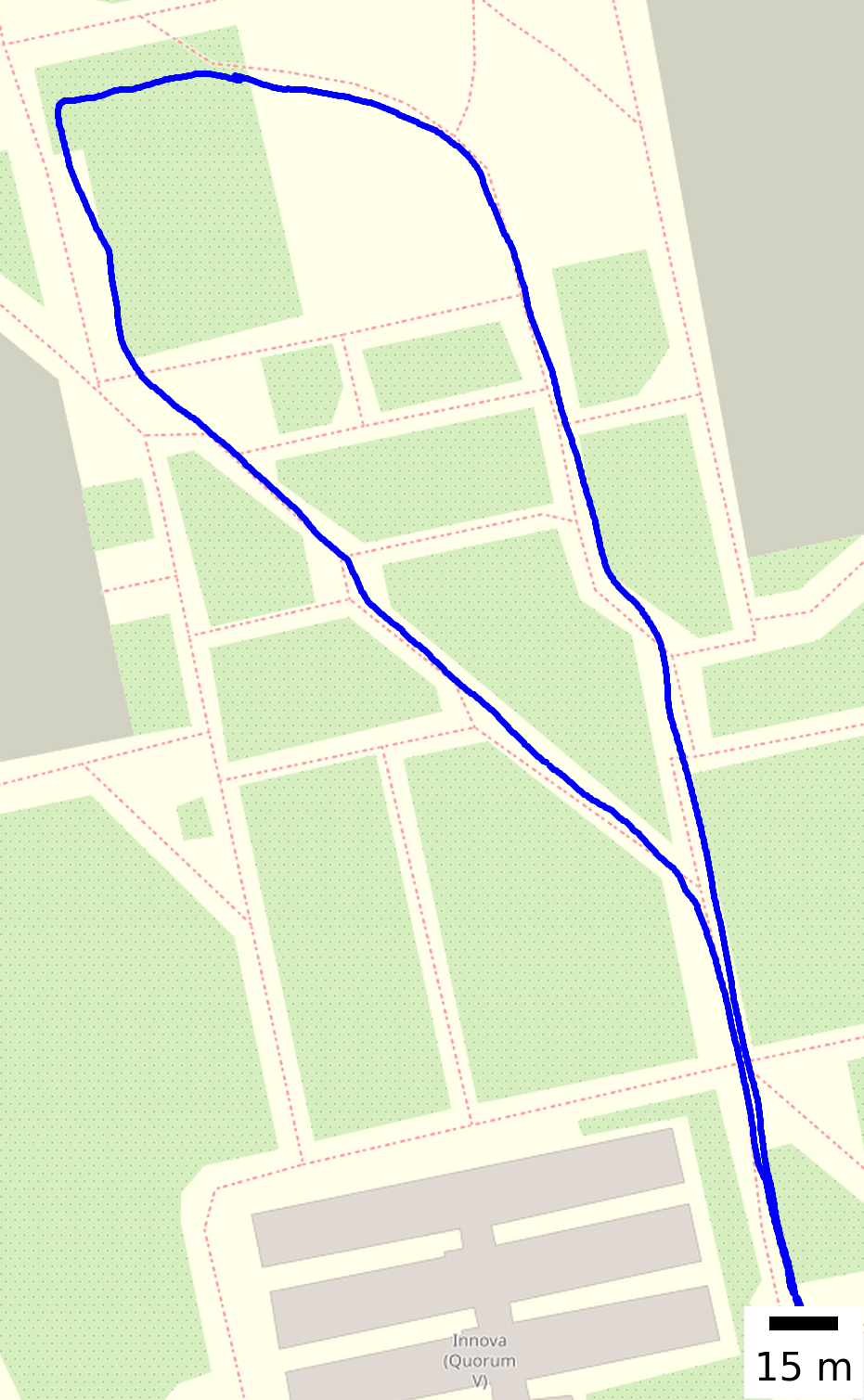}
        \caption{Route used for the validation of our method.}
        \label{arvc-val-route}
    \end{subfigure}
    \caption{ Routes on the ARVC dataset.
(a) all routes available in the dataset; (b) run used for validation purposes, plotted on OpenStreetMaps from their GPS information.}
    \label{arvc-runs}
\end{figure}

We captured several different short runs across the campus for each different season. Figure \ref{arvc-all} shows the total extension of the routes captured with our dataset across the university, while Figure \ref{arvc-val-route} shows the route selected for validation purposes (it was taken twice). 

As shown in the setup (Figure \ref{setup}), the point cloud data in this dataset was collected using an Ouster OS1-128 sensor, which operates with 128 channels and it can capture up to 2,621,440 points per scan according to its technical specifications, providing high-resolution environmental information. Two routes from this dataset were used for evaluating the descriptor, with test areas selected as arbitrary 10x10 meter regions due to the shorter route lengths in this dataset. Additionally, to reduce noise in the point clouds, points within a range of less than 2 meters or more than 60 meters from the sensor were filtered out.
\clearpage
To summarize the variety of different LiDARs used by the different datasets involved in our work, Table \ref{lidar-comparison} illustrates a comparison between each sensor per dataset. The parameters taken into account for this comparison are the different fields of view (FoV), both horizontal and vertical, which give us a general perspective of the area covered by the sensor. The table also shows the value of the points captured per second by each sensor, as this gives us an idea of how dense each point cloud is.

\begin{table}[h]
    \centering
    \begin{tabular}{c|c|c|cccc}
    \toprule
      Dataset & \makecell{Number\\of runs} & LiDAR model  &  \makecell{Horizontal FoV\\(deg.)}   &  \makecell{Vertical FoV\\(deg.)} & Channels  & \makecell{Points\\per sec.}\\
      \midrule
       Oxford & +100 & SICK LMS-151 (2D) &  270   &  (0.25, 0.5) & 1 (2D) & 50K\\
       Usyd & 60 &  Velodyne Puck VLP-16   &  360  & (+15, -15) & 16 & 0.3M \\
       KITTI &  5  &  Velodyne HDL-64E   &  360  & (+2, -24.8) & 64 & 1.3M \\
       NCLT &  27 &  Velodyne HDL-32E   &  360  & (+10.67, -30.67) & 32 & 1.3M \\
       \makecell{ARVC (under\\development)} & 15 &  Ouster OS1-128   &  360  & (+22.5, -22.5) & 128 & 2.6M  \\
       \botrule
    \end{tabular}
    \caption{Comparison of the characteristics of the different datasets involved in our work.}
    \label{lidar-comparison}
\end{table}

\subsection{Network parameters}\label{subsec42}
MinkUNeXt-SI was trained using two datasets, as mentioned in Section \ref{subsec41}: Oxford RobotCar, including intensity information, and USyd. Training was performed on an Nvidia GeForce RTX 3090 GPU, with an average training time of 60 hours for the hyperparameters listed specified in Table \ref{hyperparameters}. 

In order to find the optimal configuration for MinkUNeXt-SI, several parameter adjustments were evaluated. The different selected approaches are related to the processing of the point cloud and the velocity and computational weight of the network. We assessed different values of the quantization size, which is the size of the voxel grid used to discretize the values of the point cloud for applying the spare convolutions of our MinkUNeXt-SI. We also analyzed different values of voxel size, which refers to the size of the grid in the downsampling preprocessing to reduce the density of the point cloud before it enters the network. Finally, different values of batch size and split size were tested. The batch size refers to the number of examples that are processed simultaneously by the network, while the split size refers to the size of the parts into which the batch size is divided to facilitate the processing of large batch sizes. It has been tested that large batch sizes can lead to more stable convergence, but at the cost of high memory consumption. This network requires a large batch size for optimal performance, as it operates with the Truncated Smooth Average Precision loss function \cite{komorowski2022}.

\begin{table}[h]
\caption{Hyperparameter values for our experiments.}\label{hyperparameters}%
\begin{tabular}{@{}ll@{}}
\toprule
Parameter & Value\\
\midrule
Batch size    & 2048  \\
Epochs    & 400  \\
Split size    & 16  \\
Initial Learning Rate    & $1 \times 10^{-3}$
  \\
LR Scheduler Steps    & 250, 350  \\
L2 Weight Decay    & $1 \times 10^{-4}$
  \\
Sigmoid Temperature ($\tau$)    & 0.01  \\
\botrule
\end{tabular}
\end{table}

\begin{figure}[h]
    \centering
    \includegraphics[width=\linewidth]{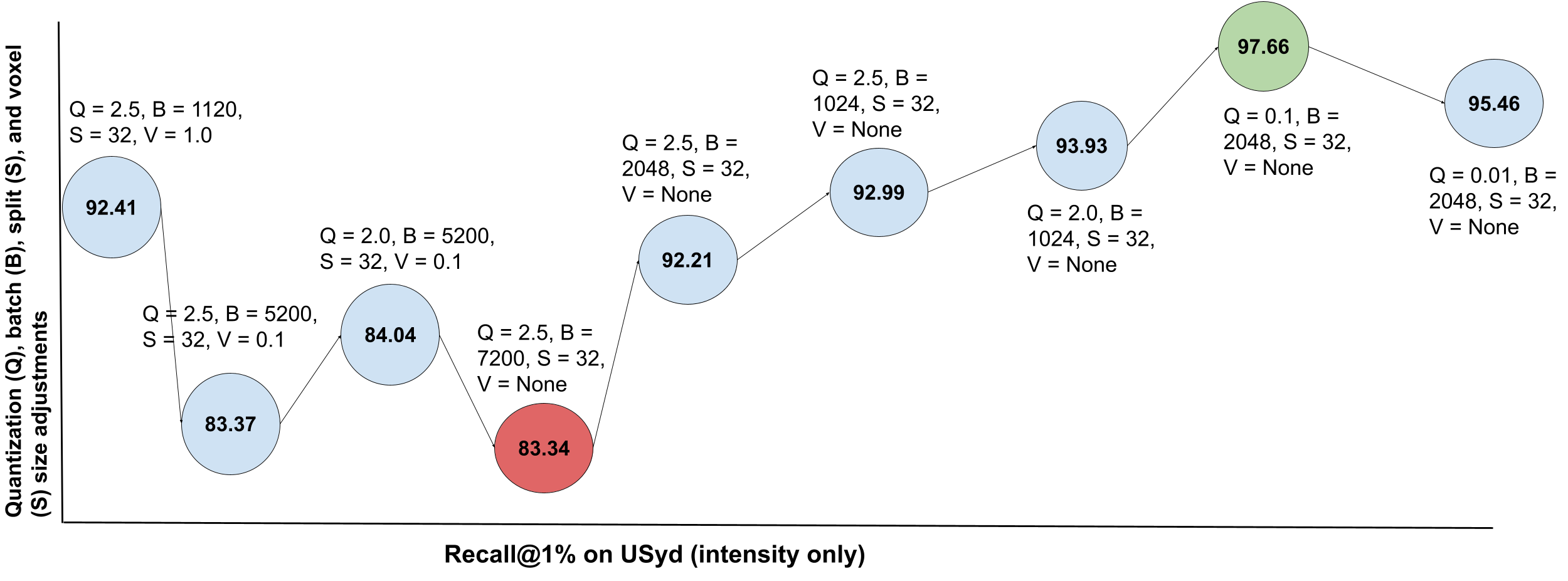}
    \caption{Graph that illustrates the design decisions for the values of the hyperparameters.}
    \label{param-graph}
\end{figure}

Regarding the design decisions made for selecting these parameters, the graph in Figure \ref{param-graph} presents different Recall@1\% results varying the value of the parameters batch size (B), batch split size (S), quantization size (Q), and voxel size (V) based on training with the USyd dataset, including intensity and Cartesian coordinates. These results indicate that an excessively large batch size yields poor performance, just as a very small one does. Additionally, it is preferable not to apply voxel downsampling as a preprocessing step before feeding the point clouds into the network, as this could lead to the loss of relevant information. In Figure \ref{param-graph}, the cases where voxel downsampling is not applied are indicated with V = None.

Regarding the quantization size, Figure \ref{param-graph} also shows that while a larger value speeds up training, it results in information loss. Moreover, this hyperparameter varies depending on the dataset used for training and whether spherical coordinates are employed. It requires the use of a spherical grid size when working with spherical coordinates. These adjustments are reflected in Table \ref{quantization}.

\begin{table}[h]
\caption{Quantization size values for our experiments.}\label{quantization}%
\begin{tabular}{@{}lll@{}}
\toprule
Train dataset & Spherical coordinates  & Value \\
\midrule
USyd    & No   & 0.1  \\
USyd    & Yes   & [0.1, 2.0, 1.875]  \\
Oxford    & No   & 0.01  \\
Oxford    & Yes   & [0.01, 2.0, 1.875]  \\
\botrule
\end{tabular}
\end{table}

To minimize the error from the loss function, this method uses the Adam optimizer with an initial learning rate of 0.01. When the specified milestones are reached, this learning rate is reduced by a factor of ten to prevent the loss function from overshooting the global minimum.

\subsection{Clustering}\label{subsec52}
\begin{figure}[h]
    \centering
    \begin{subfigure}
        [b]{0.3\textwidth}
        \includegraphics[width=\textwidth]{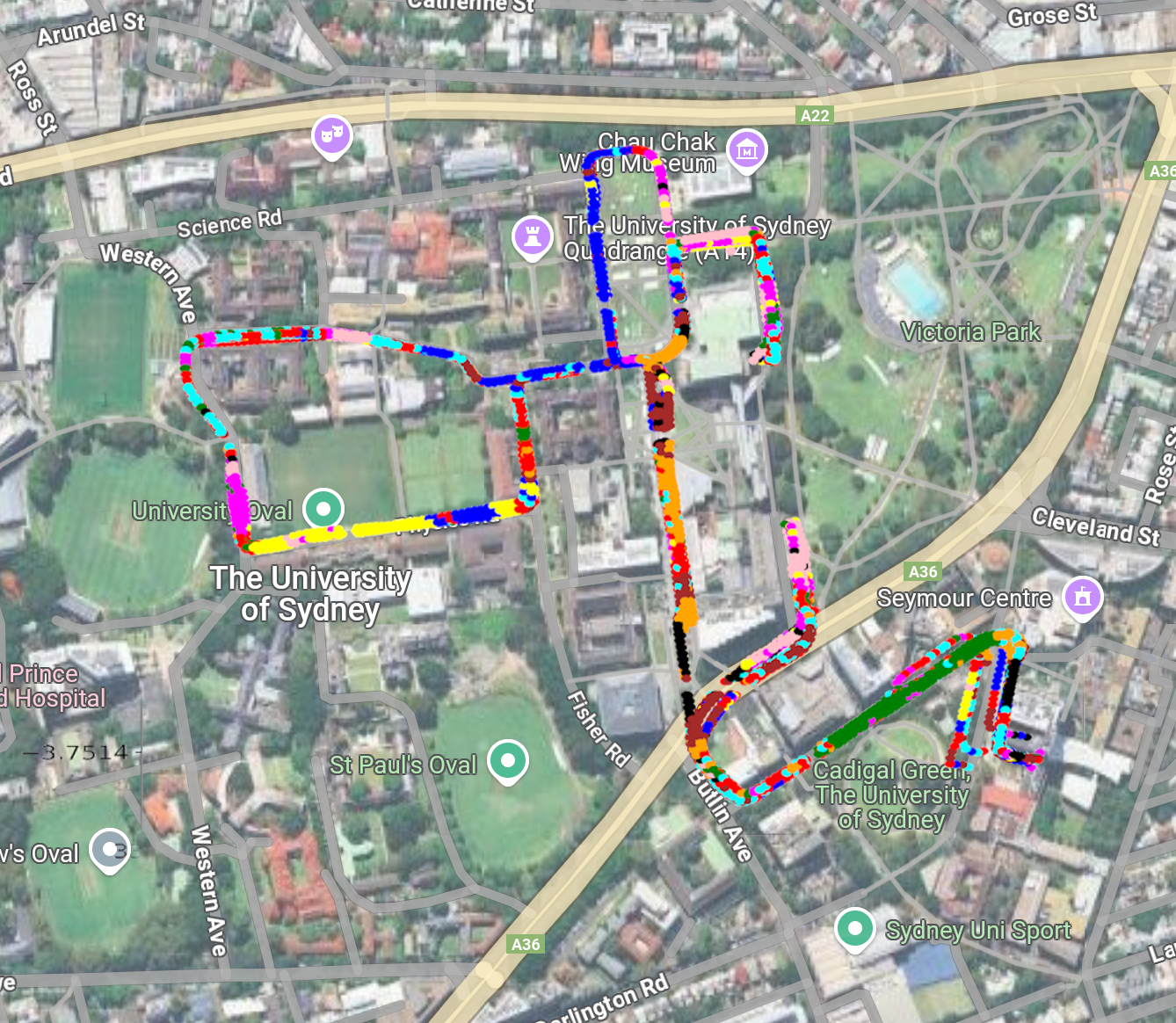}
        \caption{$k = 10$.}
        \label{clustering-usyd}
    \end{subfigure}
    \begin{subfigure}
        [b]{0.3\textwidth}
        \includegraphics[width=\textwidth]{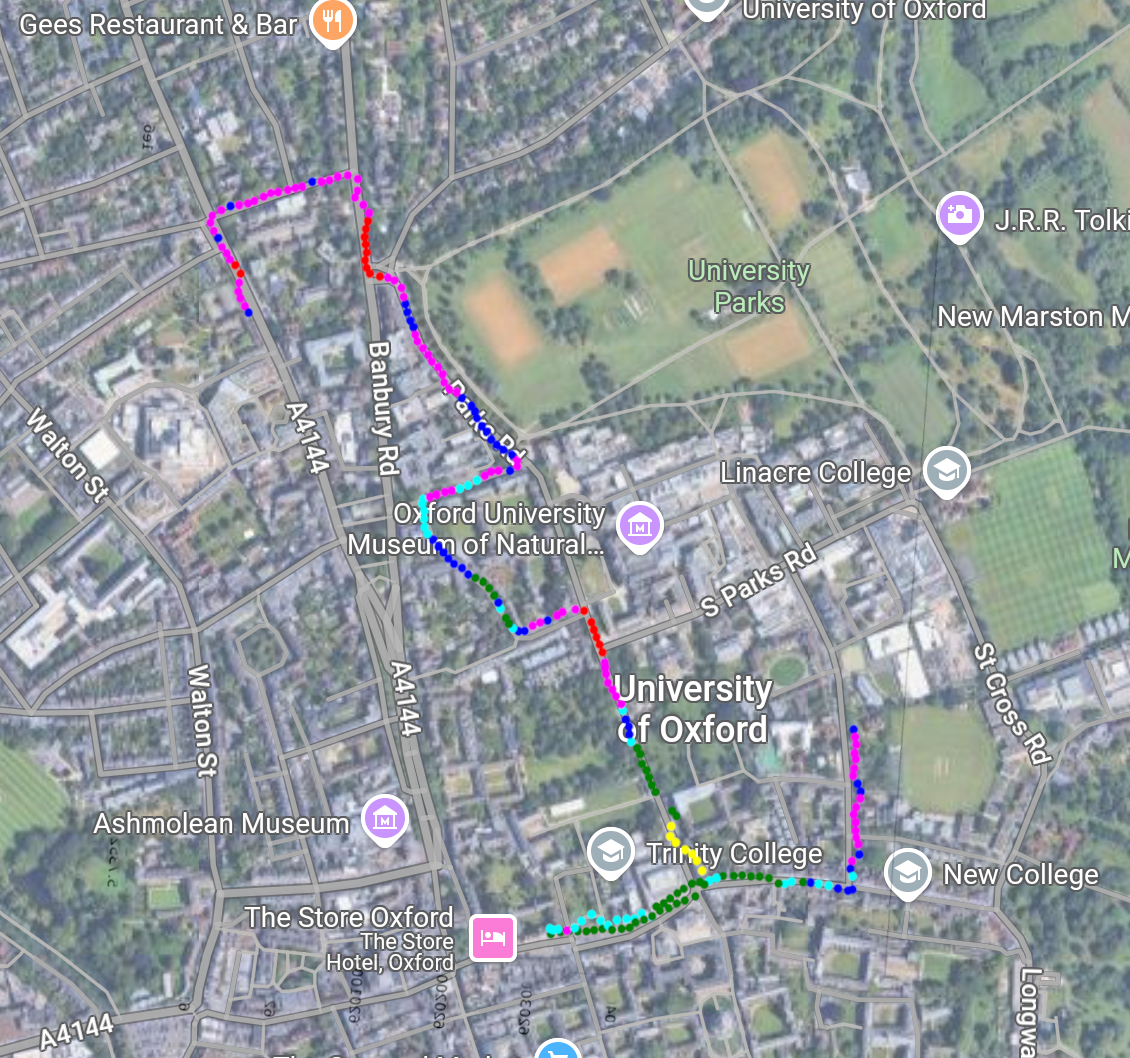}
        \caption{$k = 6$.}
        \label{clustering-oxford}
    \end{subfigure}
    \begin{subfigure}
        [b]{0.3\textwidth}
        \includegraphics[width=\textwidth]{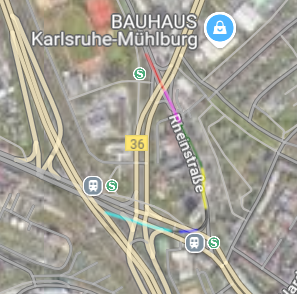}
        \caption{$k = 10$.}
        \label{clustering-kitti}
    \end{subfigure}
    \begin{subfigure}
        [b]{0.45\textwidth}
        \includegraphics[width=\textwidth]{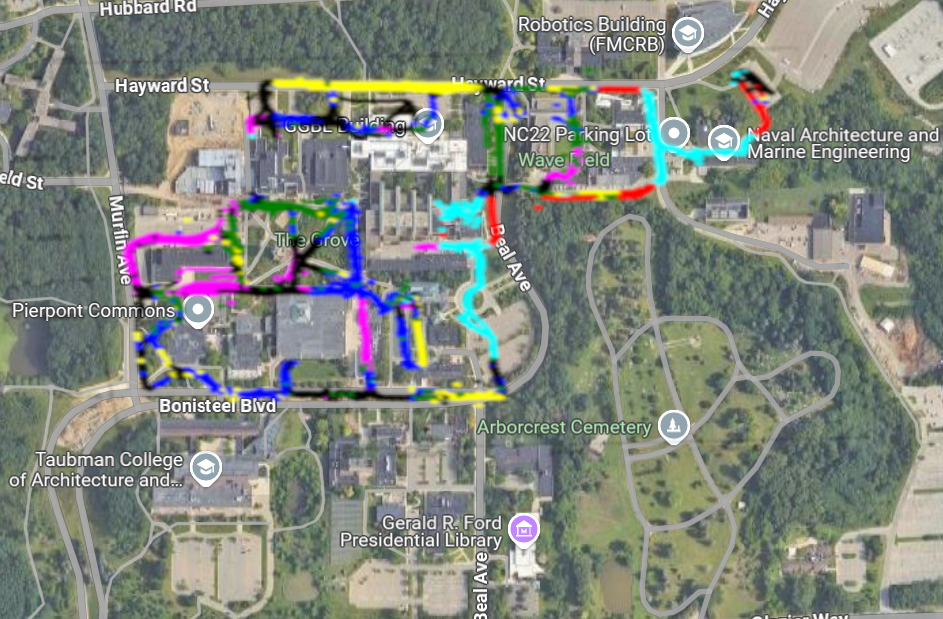}
        \caption{$k = 7$.}
        \label{clustering-nclt}
    \end{subfigure}
    \begin{subfigure}
        [b]{0.3\textwidth}
        \includegraphics[width=\textwidth]{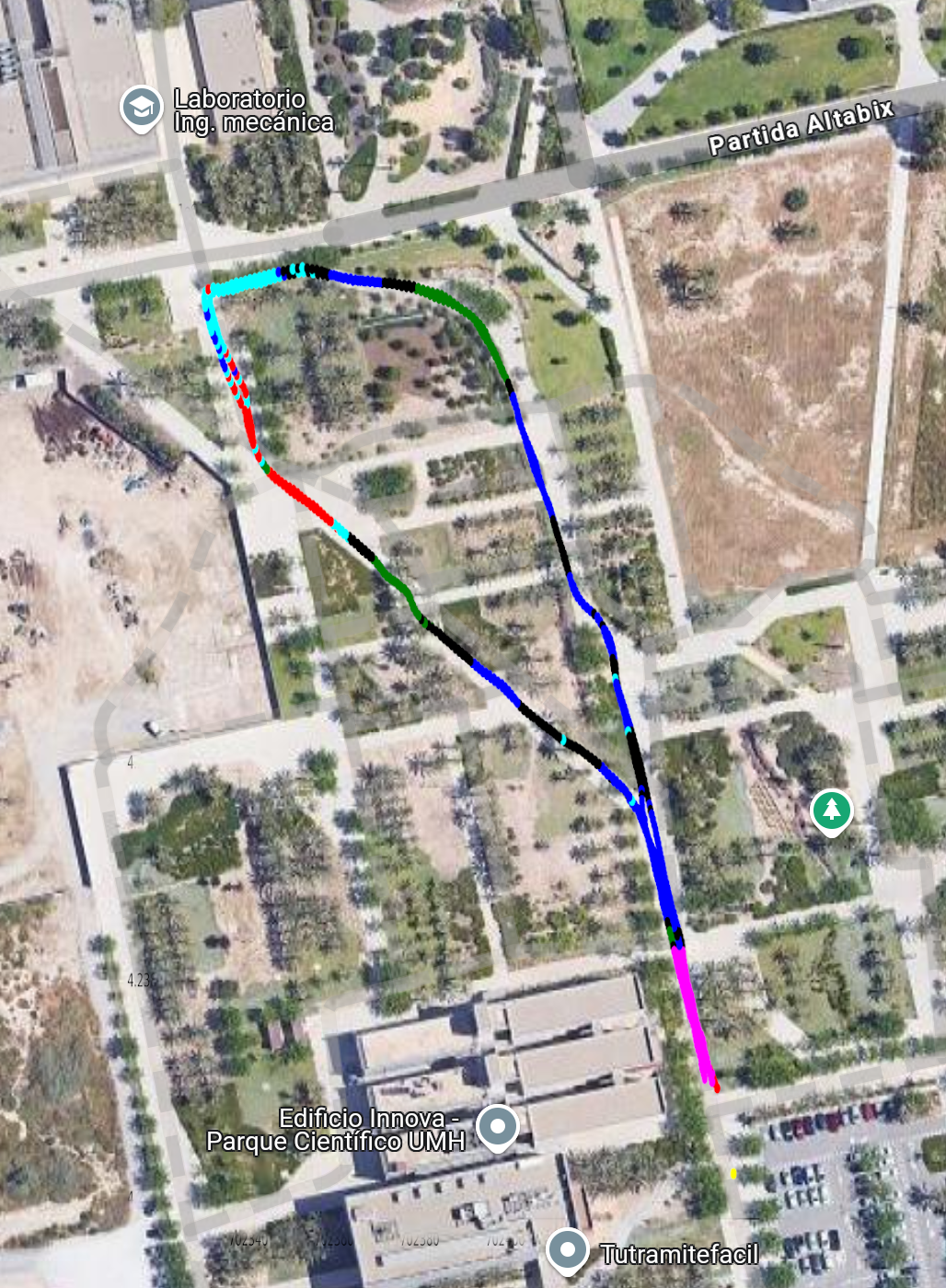}
        \caption{$k = 6$.}
        \label{clustering-arvc}
    \end{subfigure}
    
    \caption{Clustering from the descriptors obtained by training our MinkUNeXt-SI on a single run of each of the different datasets involved in our work using the k-means algorithm: (a) USyd; (b) Oxford; (c) KITTI; (d) NCLT; (e) ARVC.}
    \label{clustering-all}
\end{figure}

To verify whether our method truly generates a robust descriptor against seasonal and dynamic changes in urban environments, we computed the distance between the descriptors of the points in the evaluation sets of all datasets. The descriptor distances for all cases, as shown in Figure \ref{clustering-all}, displays the descriptor distance maps for every dataset. Concretely, it shows the different results on Figure \ref{clustering-usyd} for USyd, Figure \ref{clustering-oxford} for Oxford, Figure \ref{clustering-kitti} for KITTI, Figure \ref{clustering-nclt} for NCLT and Figure \ref{clustering-arvc} for ARVC. This clustering was obtained using the k-means algorithm \cite{ahmed2020}, selecting a total of 10 clusters on USyd, 6 clusters on Oxford, 10 for KITTI, 7 for NCLT, and 6 for ARVC. The clustering output present satisfactory
results, as the different areas that form the routes are clearly distinguishable from one
another.

The selection of the number of clusters for the k-means algorithm was done manually after careful consideration of which value would give more clarity to the presentation.

%For the KITTI dataset, the clustering process differed slightly from that of the other datasets. In the latter, the descriptor was plotted based on the UTM position associated with the point cloud from which the descriptor was obtained. However, in the case of KITTI, the descriptor was plotted using the translation values $t_x$ and $t_y$ associated with the corresponding point cloud in the given transformation matrix.
\clearpage

\section{Results}\label{sec5}
In this section the results obtained with our MinkUNeXt-SI are presented. We establish a comparison with the results obtained in \cite{zywanowski2022}, since it is the most representative work in the state-of-the-art whithin our context: a combination of using spherical coordinates and intensity as an input and processing this information with a deep learning method based on Minkowski convolutions. The code of our implementation is publicly available on a Github repository\footnote{\url{https://github.com/JudithV/MinkUNeXt-SI}} for free reproduction of these experiments.

The results obtained with our method compared to the approach in \cite{zywanowski2022} can be observed in Tables \ref{results-1percent} and \ref{results-1} for the Recall@1 and Recall@1\% values, respectively.

Regarding the results obtained in the case of training conducted with the USyd dataset, Tables \ref{results-1percent} and \ref{results-1} demonstrate that our method achieves state-of-the-art results. While it does not surpass the method to which we compare, it is capable of matching its performance by applying the same approach of transforming point cloud coordinates and incorporating intensity information. When including the intensity values for the points, the results are 97.66\% and 92.03\% for Recall@1\% and Recall@1, respectively. We obtain similar results when using the spherical representation of the coordinates without adding the intensity information for each point as a feature, obtaining results of 97.77\% and 90.36\% for Recall@1\% and Recall@1, respectively. However, we can see that the results are much better when both data information are taken into account, reaching values of 98.44\% and 92.02\% for Recall@1\% and Recall@1, respectively. The comparison tables show that the results obtained with MinkUNeXt-SI are on par with MinkLoc3D-SI \cite{zywanowski2022} and the spherical and intensity (SI) casuistic gives the best result.

In the case of the Oxford RobotCar Dataset, the improvement over the results obtained in MinkLoc3D-SI \cite{zywanowski2022} using our method is remarkable, as shown in Tables \ref{results-1percent} and \ref{results-1}. Our approach improves in all scenarios, reaching Recall@1\% values of 99.15 when using Cartesian coordinates along with intensity information. The results obtained using spherical coordinates also show a significant improvement compared to the state-of-the-art \cite{pan2021}. However, the performance decreases compared to the use of Cartesian coordinates since, in the case of this dataset, the point clouds are constructed from multiple 2D LiDARs. As a result, the spatial reconstruction lacks relevant information compared to the use of pure 3D sensors. Taking this into account, it is reasonable that the results are lower when using spherical coordinates instead of the Cartesian representation (none with intensity), with recalls of 96.28\% and 89.91\% for Recall@1\% and Recall@1, respectively. When the intensity is added and the spherical coordinates are also used (SI casuistic), the results are better than the sphericals-only casuistic, reaching values of 97.93\% and 93.22\% for Recall@1\% and Recall@1, respectively. This shows that adding the intensity values as an input to MinkUNeXt-SI improves the overall results of the solution and that these values are treated as relevant information. %This is because, if the coordinates in the spherical space are not as representative as the default Cartesian representation, including the intensity may not add relevant information to the solution. 

In general lines, these results show that our MinkUNeXt-SI reaches the state-of-the-art level of results and even it achieves better results in several scenarios if we establish a comparison with MinkLoc3D-SI \cite{zywanowski2022}, as the most similar approach to ours.

\begin{landscape}
    
\begin{table}[h]
\caption{Comparison Recall@1\% results obtained on both datasets between our method and MinkLoc3D-SI.}\label{results-1percent}%
\begin{tabular}{@{}l|ll|ll|ll|ll@{}} % anteriormente: @{}lllllll@{}
\toprule
Dataset  & MinkLoc3D-S & MinkUNeXt-S (ours) & MinkLoc3D-I & MinkUNeXt-I (ours) & MinkLoc3D-SI  & MinkUNeXt-SI (ours) \\
\midrule
USyd   & \textbf{98.8}  & 97.77  & \textbf{98.2}   & 97.66 &   \textbf{99.0}  &   98.44\\
Oxford  &  92.0  &  \textbf{96.28}  & 98.1 &  \textbf{99.15}  &  93.4   &  \textbf{97.93}  \\
\botrule
\end{tabular}
\end{table}
%\begin{landscape}
\begin{table}[h]
    \caption{Comparison Recall@1 results obtained on both datasets between our method and MinkLoc3D-SI.}\label{results-1}%
    \begin{tabular}{@{}l|ll|ll|ll|ll@{}}
        \toprule
        Dataset & MinkLoc3D-S & MinkUNeXt-S (ours) & MinkLoc3D-I  & MinkUNeXt-I (ours) & MinkLoc3D-SI & MinkUNeXt-SI (ours) \\
        \midrule
        USyd  & \textbf{93.9} &  90.36  & \textbf{92.3} & 92.03  &  \textbf{94.7}  & 92.02\\
        Oxford  &  79.9 &  \textbf{89.91}  &  93.6 &  \textbf{97.31}  &  82.2  &  \textbf{93.22} \\
        \botrule
    \end{tabular}
\end{table}
\end{landscape}
\subsection{Generalization}\label{subsec51}
Another aspect in which MinkUNeXt-SI differs from the state-of-the-art is on its generalization capability, which has been tested on several urban datasets (KITTI, NCLT and ARVC). In the case of validating our solution with the KITTI dataset \cite{geiger2013}, Table \ref{generalization-kitti} demonstrates that the generalization is better with MinkUNeXt-SI with respect to both results obtained by MinkLoc3D-SI \cite{zywanowski2022}. Again, this outcome reveals improvement over the state-of-the-art, reaching values Recall@1\% values of 84.14\% and a Recall@1 of 83.33\% in the case of having trained our method on USyd, and Recall@1\% results of 82.14\% and Recall@1 of 72.62\% using our solution trained on the Oxford RobotCar dataset. Furthermore, if we take a look at the previous method validation results on this dataset, we can see that MinkUNeXt-SI achieves an outstanding improvement in this area.

\begin{table}[h]
\caption{Generalization results obtained on the KITTI dataset compared to other state-of-the-art methods.}\label{generalization-kitti}% compared to other state-of-the-art methods.
\begin{tabular}{@{}lllll@{}}
\toprule
Method & Trained on & Source of results &   Recall@1\% & Recall@1\\
\midrule
PointNetVLAD \cite{uy2018}  &    Oxford &  \cite{pan2021} &  72.43  & -  \\
MinkLoc3D \cite{komorowski2021}  &    USyd &  \cite{zywanowski2022} &  73.8  & 69.1  \\
LPD-Net \cite{liu2019}  &    Oxford &  \cite{pan2021} &  74.58  & -  \\
AugNet \cite{oertel2020}  &    Oxford &  \cite{pan2021} &  75.60  & -  \\
Coral-VLAD \cite{pan2021}  &    Oxford &  \cite{pan2021} &  76.43  & -  \\
MinkLoc3D-SI    &    USyd &  \cite{zywanowski2022} &  81.0  & 78.6  \\
MinkLoc3D-SI    &    Oxford &  \cite{zywanowski2022} &  81.0  & 72.6  \\
MinKUNeXt-SI (ours)    &    USyd & our evaluation    &  \textbf{84.14}  & \textbf{83.33}  \\
MinKUNeXt-SI (ours)    &    Oxford & our evaluation  &  \textbf{82.14}  & \textbf{72.62}  \\
\botrule
\end{tabular}
\end{table}

%previous table: other source of results:
%PN-Vlad \cite{uy2018}: 72.43
%LPD-Net \cite{liu2019}: 74.58
%AugNet: 75.60
%CORAL-VLAD \cite{pan2021}: 76.43
%Anteriores cuatro resultados: paper CORAL: Colored structural representation for bi-modal place recognition
%MinkLoc++ \cite{komorowski2021minkloc++}: 75.0
%MinkLoc3D: 73.8
%Anterior resultado extraido tb de MinkLoc3d-SI

The problem with the KITTI dataset for our purpose is that it was not collected using a long-term approach. To overcome this limitation, we also evaluated our method using the NCLT and ARVC datasets. The approach of using NCLT as a validation source is a novel approach in the state-of-the-art, since its use case was for training and testing, not as an unseen scenario \cite{pan2021}.

\begin{table}[h]
\caption{Generalization results obtained on the NCLT and ARVC datasets by MinkUNeXt-SI.}\label{generalization-others}%
\begin{tabular}{@{}llll@{}}
\toprule
Trained on & Evaluated on  &  Recall@1\% & Recall@1\\
\midrule
USyd    &    NCLT   &  88.84  & 61.49  \\
USyd    &    ARVC    &  97.98  & 88.59  \\
Oxford    &    NCLT   &  68.88  & 33.61  \\
Oxford    &    ARVC  &  94.62  & 87.90  \\
\botrule
\end{tabular}
\end{table}

Table \ref{generalization-others} shows the results after evaluating MinkUNeXt-SI on these datasets. The results are again outstanding, outperforming the state-of-the-art with Recall@1\% values of 88.84\% when generalizing to NCLT and 97.98\% in the case of ARVC over our USyd-trained solution. In the location recognition task, when comparing results and validating over different datasets, the LiDAR sensor that captured the raw information in each of these collections is of great importance. Taking this into account, we can observe that in the case of training our method with the Oxford dataset, the value of both recalls is lower due to the significant differences between the LiDARs used, since the scans of the Oxford dataset are built from 2D LiDARs, while the other datasets use 3D LiDARs, which provide sparser information. Again, the recall results in this scenario are exceptional in the case of the ARVC dataset with values of 94.62\% and 87.90\% for Recall@1\% and Recall@1, respectively. With these results, we can also observe that the high quality of the LiDAR and the precision of the GPS RTK sensor of the ARVC data set allow for a better success of the generalization than with the other datasets.

\subsection{Inference time}\label{subsec53}
\begin{figure}[h]
    \centering
    \includegraphics[width=0.7\linewidth]{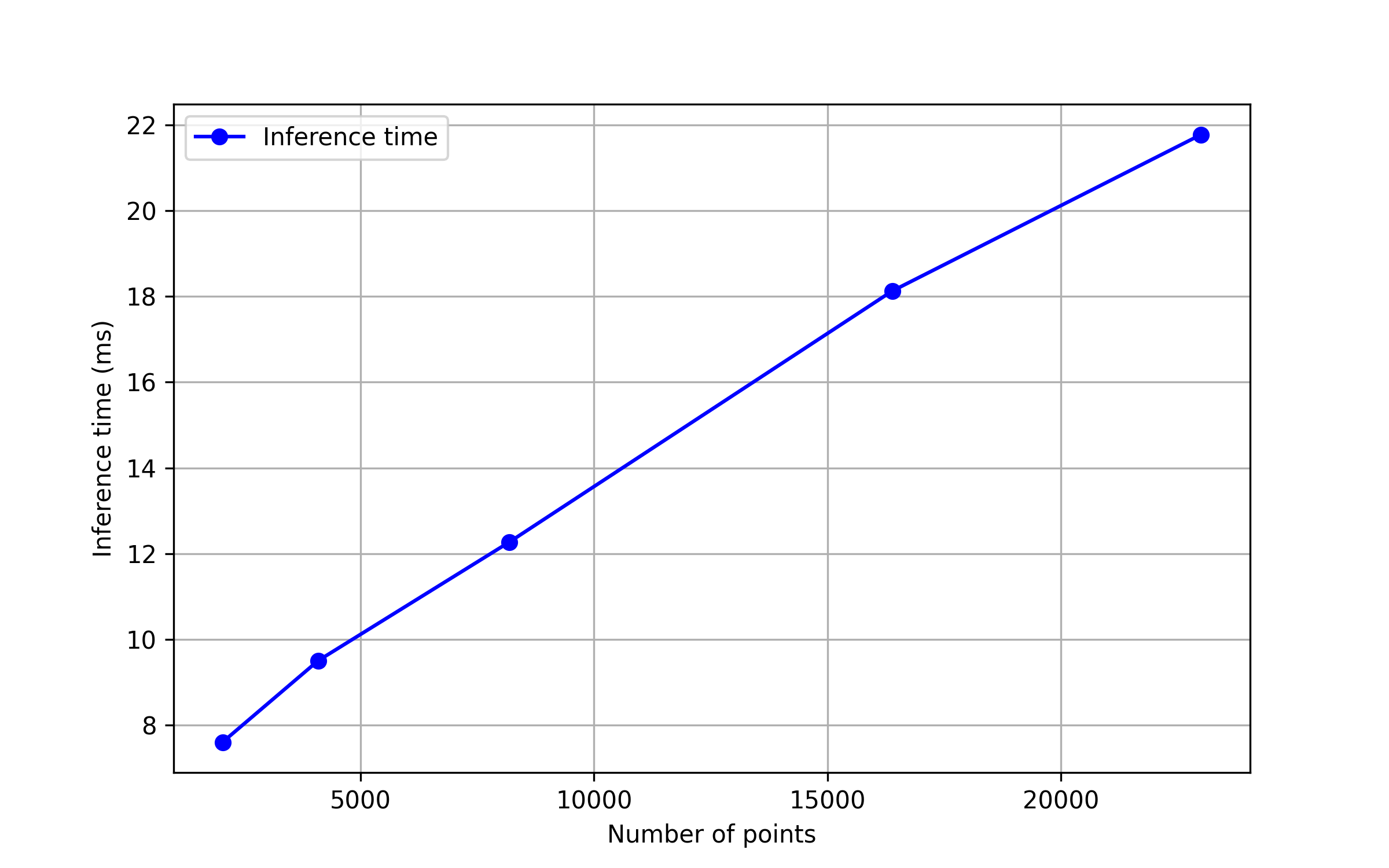}
    \caption{Inference time for one point cloud with our method (MinkUNeXt-SI).}
    \label{inference}
\end{figure}
The number of points present in a point cloud depends on different factors: the sensor's resolution, which is given by the number of channels, as well as the frequency rate of such sensor. It is evident that the number of points differs significantly in the case of the Oxford dataset (which was collected more than ten years ago and constructs point clouds from 2D LiDARs) and the resolution provided by the ARVC dataset, which was captured using an Ouster OS1-128 sensor. This causes the network to process point clouds with highly varying resolutions, ranging from 4,096 points in the Oxford dataset to up to 25,000 points in the USyd dataset, where the higher-resolution clouds have been processed to remove noise by filtering based on minimum and maximum radius. Figure \ref{inference} shows that while inference time increases as the point cloud size grows, the computation remains very fast, with a value of less than 25 ms for 23,000 points. MinkUNeXt-SI's viability in a real-time operating environment is ensured by its fast response time.

\section{Conclusion}\label{sec6}
This paper proposes the inclusion of intensity information along with spherical space coordinates to enhance place recognition descriptors obtained through the MinkUNeXt network, ensuring robustness against dynamic and seasonal changes. Additionally, histogram equalization is introduced as a normalization method for intensity values within a 0 to 1 range, providing greater structural information about the scene. The use of a descriptor of the environment instead of raw data is highly beneficial for the place recognition task, due to its compact design it results in a powerful tool that is computationally efficient, facilitating the process of global search for real-time localization on mobile robots or autonomous driving systems.

The proposed method demonstrates that incorporating spherical coordinates and intensity information is highly beneficial for place recognition, as it achieves excellent results in both training and evaluation, reaching in some extents and clearly surpassing in others the current state-of-the-art. Results are consistently improved when the spherical transformation and intensity information for each point is included, compared to training without it. This was demonstrated in the Section \ref{sec5}, which presented results that surpassed the state-of-the-art on this area. In addition, as seen in Section \ref{subsec51}, MinkUNeXt-SI achieves excellent generalization results in several different environments that include weather changes, dynamic elements, and seasonal changes in the case of long-term datasets. This demonstrates that our descriptor is robust against all kinds of challenging situations and it can be tested in any environment. The code of MinkUNeXt-SI is publicly available for use and reproducibility of results.

As future work, we consider the fusion of LiDAR data from different sources in the training process, aiming to obtain a descriptor that generalizes well across various environments, regardless of the sensor model used to capture the point clouds in different datasets.

\backmatter
\bmhead{Availability of data and material}
All data used for the development of this work, including the method's code and the datasets used for the experiments, are publicly available on Github.

\bmhead{Competing interests}

The authors declare that they have no known competing financial interests or personal relationships that could have appeared to influence the work reported in this paper.

\bmhead{Funding}
This research work is part of the project TED2021-130901B-I00 funded by MICIU/AEI/10.13039/501100011033 and by the European Union NextGenerationEU/PRTR. It is also part of the project PID2023-149575OB-I00 funded by MICIU/AEI/10.13039/501100011033 and by FEDER, UE.

\bmhead{Authors' contributions}

Conceptualization: Judith Vilella-Cantos, Mónica Ballesta; Methodology: Judith Vilella-Cantos, David Valiente; Software: Judith Vilella-Cantos, Juan José Cabrera; Validation: Judith Vilella-Cantos, Juan José Cabrera, David Valiente, Luis Payá; Formal analysis: Judith Vilella-Cantos, Mónica Ballesta, David Valiente; Investigation: Judith Vilella-Cantos; Resources: Judith Vilella-Cantos; Data curation: Judith Vilella-Cantos; Writing—original draft preparation: Judith Vilella-Cantos; Writing—review and editing: David Valiente, Luis Payá, Mónica Ballesta; Visualization: Judith Vilella-Cantos; Supervision: Mónica Ballesta, David Valiente, Luis Payá; Project administration: Luis Payá; Funding acquisition: Luis Payá. All authors have read and agreed to the published version of the manuscript.

\bmhead{Acknowledgements}

This research work has been made thank to the funding of the project TED2021-130901B-I00. This project was funded by MICIU/AEI/10.13039/501100011033 and by the European Union NextGenerationEU/PRTR. It is also part of the project PID2023-149575OB-I00 funded by MICIU/AEI/10.13039/501100011033 and by FEDER, UE.


\bibliography{sn-bibliography}

\end{document}